\pdfoutput=1
\documentclass[11pt]{article}

\usepackage{booktabs}
\usepackage{caption}
\usepackage{graphicx}

\usepackage[final]{coling}

\usepackage{times}
\usepackage{latexsym}

\usepackage[T1]{fontenc}
\usepackage[utf8]{inputenc}

\usepackage{microtype}

\usepackage{inconsolata}

\usepackage{graphicx}

\title{Adapting Multilingual LLMs to Low-Resource Languages using Continued Pre-training and Synthetic Corpus}

\author{Raviraj Joshi, Kanishk Singla, Anusha Kamath, Raunak Kalani, Rakesh Paul, \\ \textbf{ Utkarsh Vaidya, Sanjay Singh Chauhan, Niranjan Wartikar, Eileen Long} \\
        NVIDIA \\
        \texttt{\{ravirajj, kanishks, anushak, rkalani, rapaul, uvaidya} \\ \texttt{schauhan, nwartikar, elong\}@nvidia.com}}

\begin{document}
\maketitle
\begin{abstract}
Multilingual LLMs support a variety of languages; however, their performance is suboptimal for low-resource languages. In this work, we emphasize the importance of continued pre-training of multilingual LLMs and the use of translation-based synthetic pre-training corpora for improving LLMs in low-resource languages. We conduct our study in the context of the low-resource Indic language Hindi. We introduce Nemotron-Mini-Hindi 4B, a bilingual SLM supporting both Hindi and English, based on Nemotron-Mini 4B. The model is trained using a mix of real and synthetic Hindi + English tokens, with continuous pre-training performed on 400B tokens. We demonstrate that both the base and instruct models achieve state-of-the-art results on Hindi benchmarks while remaining competitive on English tasks. Additionally, we observe that the continued pre-training approach enhances the model's overall factual accuracy. We perform an ablation study to highlight the impact of Hindi pre-training, showing significant improvements in Hindi chat capabilities and factual accuracy, which cannot be achieved through Hindi alignment alone.
%the continued pretraining improves the factuality of the model
\end{abstract}

\begin{table*}[h!]
    \centering
    \small
        \begin{tabular}{lcccccc} \hline
             {\bf Model} & {\bf Layers} & {\bf Hidden Size} & {\bf Att.~Heads} & {\bf Query Groups} & {\bf MLP Hidden} & {\bf Parameters}\\ \hline
        Nemotron 4B & 32 & 3072 & 24 & 8 & 9216 & 4.19B \\ \hline
        \end{tabular}
    \caption{Architecture details of Nemotron-Mini-4B model.}
    \label{tab:architecture}
\end{table*}

\section{Introduction}
The accuracy and utility of large language models (LLMs) have continuously improved over time. Both closed and open-source LLMs have demonstrated strong performance in English and several other languages. Open models such as Nemotron \cite{adler2024nemotron}, Gemma \cite{team2024gemma}, and Llama \cite{dubey2024llama} are inherently multilingual. For instance, the Nemotron-4 15B model was pre-trained on 8 trillion tokens, of which 15\% were multilingual \cite{parmar2024nemotron}. However, the proportion of multilingual data is limited, which in turn affects the accuracy of these models on non-English languages. 

\begin{figure}[]  
    \centering
    \includegraphics[width=\columnwidth]{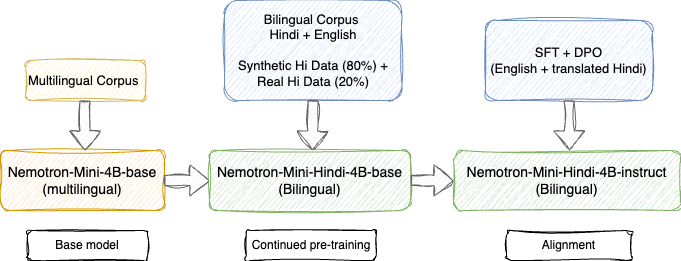}  
    \caption{Adaptation of multilingual Nemotron-Mini-4B model (also known as Minitron-4B).}
    \label{fig:minitron-process}
\end{figure}

The model's performance further diminishes as we move from high-resource to low-resource languages. In this work, we specifically focus on the Indic language Hindi as our target low-resource language. Out of the 8 trillion tokens used to train the Nemotron-4 models, only 20 billion tokens are in Hindi. As a result, while the model can understand and generate Hindi content to a reasonable extent, the usability of such a multilingual LLM for specific low-resource languages remains questionable. Frequent hallucinations, meaningless sentences, and mixing of English content often occur when responding to purely Hindi queries in the Devanagari script. There is a strong need to adapt multilingual LLMs to target languages to enhance their usability.

Recently, in the context of Indic languages, target language Supervised Fine-Tuning (SFT) has become a common practice to adapt LLMs to specific languages \cite{gala2024airavata}. However, it remains to be studied whether language-specific SFT tuning improves LLMs' understanding in regional contexts. Some studies suggest that SFT can introduce LLMs to new domain knowledge, though it is typically used to enhance the model’s instruction-following capability \cite{mecklenburg2024injecting}. SFT on translated English instruction tuning data is widely used to develop regional LLMs for Indic languages. While this may improve instruction-following in the target language, it may not enhance LLMs' understanding of regional contexts \cite{balachandran2023tamil}. Another approach to updating LLM knowledge is continued pre-training, but the limited availability of tokens for low-resource languages makes this both infeasible and prone to overfitting.

In this work, we focus on a continued pre-training approach using a mix of real and synthetic corpora. We demonstrate that a robust base model can be adapted to the target language with a small continued pre-training corpus. This approach is particularly relevant for low-resource languages, where the amount of training data is limited. The synthetic pre-training dataset is curated by translating high-quality generic English corpora into the target language. To further expand the corpus and support Roman script queries in the target language, the text is transliterated into Roman script and used for pre-training. The base model is then aligned using supervised fine-tuning (SFT), followed by preference tuning with Direct Preference Optimization (DPO). We observe that the continued pre-training approach is particularly useful for reducing hallucinations, improving regional knowledge of LLMs, and enhancing response capabilities in the target language. The high-level process is outlined in Figure \ref{fig:minitron-process},

Based on this approach, we present Nemotron-Mini-Hindi-4B-Base\footnote{\url{https://huggingface.co/nvidia/Nemotron-4-Mini-Hindi-4B-Base}} and Nemotron-Mini-Hindi-4B-Instruct\footnote{\url{https://huggingface.co/nvidia/Nemotron-4-Mini-Hindi-4B-Instruct}}\footnote{\url{https://build.nvidia.com/nvidia/nemotron-4-mini-hindi-4b-instruct}}, state-of-the-art Small Language Models (SLMs) for the Hindi language. These SLMs support Hindi, English, and Hinglish. The Hindi models are based on the multilingual Nemotron-Mini-4B (also known as Minitron-4B), adapted with continued pre-training on 400 billion Hindi and English tokens. The data blend used equal proportions of both languages. The instruct version of the model was developed using SFT and DPO techniques. The model outperforms all similarly sized models on various IndicXTREME, IndicNLG benchmark tasks and popular translated English benchmarks such as MMLU, Hellaswag, ARC-C, and ARC-E \cite{gala2024airavata}. We also perform LLM-based evaluations using the benchmark datasets IndicQuest \cite{rohera2024l3cube} and in-house SubjectiveEval, with GPT-4 serving as the judge LLM. This is the first study to present and evaluate bilingual language models of this nature. We provide a thorough study of the models in both languages. 

Additionally, we perform an extensive ablation to analyze the impact of Hindi SFT and DPO on both the base multilingual Nemotron-Mini-4B and the Hindi-specific Nemotron-Mini-Hindi-4B. Our results show that while Hindi alignment improves performance, Hindi pre-training is essential for achieving strong results on Hindi benchmarks. Notably, Nemotron-Mini-Hindi 4B leads to significant gains in factual accuracy, not just for Hindi but also for English, showcasing effective cross-lingual transfer.

\section{Related Work}
In this section, we review various approaches for adapting LLMs to different languages. Several efforts have focused on adapting LLaMA models to Indic languages. A common method involves extending the vocabulary, followed by SFT or PEFT (LoRA) using translated and available SFT corpora in Indic languages. Examples of such work include OpenHathi, Airavata \cite{gala2024airavata}, Tamil-LLaMA \cite{balachandran2023tamil}, Navarasa\footnote{https://huggingface.co/Telugu-LLM-Labs/Indic-gemma-7b-finetuned-sft-Navarasa-2.0}, Ambari, MalayaLLM, and Marathi-Gemma \cite{joshi2022l3cube_marathi}. Notably, some of these efforts employ bilingual next-word prediction, alternating between English and the target language in the pre-training corpus. Airavata also introduced an evaluation framework\footnote{https://github.com/AI4Bharat/IndicInstruct} for Indic LLMs, which we leverage to evaluate Nemotron-Mini-Hindi 4B and other multilingual models.

Apart from Indic languages, similar efforts have been made for other languages, including Chinese LLaMA \cite{cui2023efficient}, LLaMATurk \cite{toraman2024llamaturk}, FinGPT \cite{luukkonen2023fingpt}, and RedWhale \cite{vo2024redwhale} for Chinese, Turkish, Finnish, and Korean, respectively. These LLMs use one or more techniques such as tokenizer extension, secondary pretraining, and supervised fine-tuning. The key distinction of our work lies in its emphasis on developing bilingual LLMs, whereas the aforementioned efforts concentrate on creating monolingual LLMs.

\citet{cahyawijaya2024llms} show that large language models can learn low-resource languages effectively using in-context learning and few-shot examples, improving performance through cross-lingual contexts without extensive tuning.
\citet{gurgurov2024adapting} enhance multilingual LLMs for low-resource languages by using adapters with data from ConceptNet, boosting performance in sentiment analysis and named entity recognition.

% Alternatively, the work by \citet{cahyawijaya2024llms} demonstrates how large language models can effectively learn low-resource languages through in-context learning with few-shot examples. It highlights the use of cross-lingual contexts and semantic relevance to improve performance without extensive parameter tuning.

% In \citet{gurgurov2024adapting}, multilingual LLMs are enhanced for low-resource languages by integrating knowledge from linguistic ontologies using adapters. This approach leverages language-specific adapters fine-tuned on data from ConceptNet, improving performance in tasks such as sentiment analysis and named entity recognition.

\section{Methodology}
In this section, we describe our methodology for adapting multilingual LLMs to target languages to improve performance in those languages. Specifically, we build a bilingual SLM that supports both Hindi and English. We conduct our adaptation experiments using the multilingual Nemotron-Mini-4B model (also known as Minitron-4B). The model undergoes continuous pre-training with an equal mixture of Hindi and English data, consisting of 200B tokens per language. The original Nemotron-4B model was primarily trained on English tokens and had seen only 20B Hindi tokens. Given the limited amount of Hindi data, adapting an existing multilingual model rather than training from scratch is an effective strategy, allowing us to leverage the knowledge learned from the pre-trained model. Additionally, as Nemotron-4B employs a large 256k tokenizer, we did not need to extend the tokenizer. The fertility ratio for Hindi text is 1.7, which is better than that of its Llama (2.64) and Gemma (1.98) counterparts.

\subsection{Synthetic Data Curation}
One of the key aspects of our work is the creation of a synthetic Hindi pre-training dataset. This synthetic data is generated using machine translation and transliteration. We first select high-quality English data sources and translate them into Hindi using a custom document translation pipeline. This pipeline preserves the document structure, including elements like bullet points and tables, and employs the IndicTrans2 model \cite{galaindictrans2} for sentence translation. However, since the translated data may contain noise, we use an n-gram language model to filter out low-quality samples. This model, trained on MuRIL-tokenized \cite{khanuja2021muril} real Hindi data, applies perplexity scores to identify and exclude noisy translations. Around 2\% of the documents were discarded post-filtering.

The translated Hindi data comprises approximately 60 billion tokens. We then combine this synthetic data with around 40 billion real tokens (web-scraped data) to create a dataset totaling 100 billion Hindi tokens. Additionally, this entire Hindi text is transliterated into Roman script, expanding the total dataset to 220 billion tokens. The transliterated tokens are included to enable the model to support Hinglish queries. This Hindi data is further combined with 200 billion English tokens for continued pre-training. Including the English dataset helps prevent catastrophic forgetting of English capabilities and contributes to training stability. Fuzzy deduplication is performed on the entire text using NeMo-Curator\footnote{https://github.com/NVIDIA/NeMo-Curator} to eliminate similar documents. The real Hindi data sources include internal web-based datasets and Sangraha Corpus \cite{khan2024indicllmsuite}. The English dataset is a subset of the pre-training corpus used for the Nemotron-15B model. All the datasets used in this work are commercially friendly.

\subsection{Continued Pre-training}
The Nemotron-Mini-4B base model is used for continuous pre-training, and its architecture details are presented in Table \ref{tab:architecture}. 
The Nemotron-Mini-4B model is derived from the Nemotron-15B model using compression techniques such as pruning and distillation, consisting of 2.6B trainable parameters \cite{muralidharan2024compact}.
Re-training is performed using a standard causal modeling objective. The dataset consists of 400B tokens, with an equal mix of Hindi and English. During batch sampling, greater weight is given to real data compared to synthetic data. We use the same optimizer settings and data split as \cite{parmar2024nemotron}, with a cosine learning rate decay schedule from 2e-4 to 4.5e-7. This model is referred to as Nemotron-Mini-Hindi-4B, a base model where Hindi is the primary language. The re-training was performed using the Megatron-LM library \cite{shoeybi2020megatronlm} and 128 Nvidia A100 GPUs.

%% base model table
\begin{table*}[h!]
    \centering
    \resizebox{\textwidth}{!}{
        \begin{tabular}{lcccccccc}
\toprule
\textbf{Base models} & \textbf{Metric} & \textbf{Nemotron-Mini-Hindi-4B} & \textbf{Nemotron-Mini-4B} & \textbf{Sarvam-1 2B} & \textbf{Gemma 2-2B} & \textbf{Openhathi} & \textbf{Llama-3.1 8B} & \textbf{Gemma 2-9B} \\ \hline
\midrule
IndicSentiment & F1 - NLU & 84.31 & 72.47 & 96.36 & 91.90 & 72.89 & 92.06 & 94.90 \\
IndicCopa & F1 - NLU & 81.86 & 62.50 & 51.63 & 58.65 & 68.69 & 61.87 & 72.58 \\
IndicXNLI & F1 - NLU & 49.67 & 40.39 & 36.08 & 16.67 & 16.67 & 16.67 & 16.79 \\
IndicXParaphrase & F1 - NLU & 37.09 & 16.27 & 80.99 & 26.60 & 71.72 & 72.75 & 71.38 \\
% Indic QA (No Context) 1 shot & F1 - NLG & 8.02 & 1.82 & 1.21 & 13.09 & 17 & 13.30 & 21.52 \\
Indic QA (With Context) 1 shot & F1 - NLG & 18.32 & 15.10 & 35.81 & 33.37 & 20.69 & 35.92 & 46.27 \\
Indic Headline 1 shot & BLEURT - NLG & 0.50 & 0.46 & 0.36 & 0.27 & 0.47 & 0.38 & 0.27 \\
IndicWikiBio 1 shot & BLEURT - NLG & 0.62 & 0.59 & 0.53 & 0.60 & 0.52 & 0.60 & 0.63 \\ \hline
MMLU & Acc - NLU & 49.89 & 38.20 & 45.65 & 35.05 & 32.27 & 44.84 & 55.08 \\
BoolQ & Acc - NLU & 71.71 & 70.79 & 56.08 & 66.00 & 58.56 & 61.00 & 61.00 \\
ARC Easy & Acc - NLU & 78.81 & 58.25 & 76.85 & 52.31 & 44.28 & 67.05 & 85.69 \\
Arc Challenge & Acc - NLU & 65.02 & 47.87 & 59.04 & 40.78 & 32.68 & 54.10 & 76.02 \\
Hella Swag & Acc - NLU & 31.66 & 25.31 & 37.13 & 27.50 & 25.59 & 33.50 & 42.40 \\

\bottomrule
\end{tabular}
        }
    \caption{Performance metrics for various base models across different Hindi tasks. The results are zero-shot unless otherwise specified.}
    \label{tab:model_performance_base}
\end{table*}

%%%% Chat model table
\begin{table*}[h!]
    \centering
    \resizebox{\textwidth}{!}{
        \begin{tabular}{lccccccccc}
\toprule
\textbf{Instruct models} & \textbf{Metric} & \textbf{Nemotron-Mini-Hindi-4B} & \textbf{Nemotron-Mini-4B} & \textbf{Airavata} & \textbf{Navarasa 2B} & \textbf{Gemma-2 2B} & \textbf{Navarasa 7B} & \textbf{Llama-3.1 8B} & \textbf{Gemma-2 9B} \\ \hline
\midrule
IndicSentiment        & F1 - NLU     & 97.62   & 90.01   & 95.81   & 93.62   & 94.32   & 95.99   & 98.59   & 99.09   \\ %\hline
IndicCopa             & F1 - NLU     & 80.1    & 66.01   & 63.75   & 38.83   & 27.64   & 62.59   & 59.08   & 89.89   \\ %\hline
IndicXNLI             & F1 - NLU     & 53.77   & 39.25   & 73.26   & 16.67   & 17.33   & 38.19   & 31.27   & 39.71   \\ %\hline
IndicXParaphrase      & F1 - NLU     & 67.93   & 83.74   & 76.53   & 43.82   & 43.06   & 44.58   & 77.72   & 61.38   \\ %\hline
Indic QA (With Context) 1 shot & F1 - NLG & 37.51   & 42.56    & 37.69   & 3.3     & 62.95   & 19.09   & 40.03   & 59.83   \\ %\hline
Indic Headline 1 shot   & BLEURT - NLG & 0.44    & 0.18    & 0.38    & 0.24    & 0.39    & 0.3     & 0.26    & 0.25    \\ %\hline
IndicWikiBio 1 shot   & BLEURT - NLG & 0.6     & 0.49    & 0.43    & 0.3     & 0.49    & 0.45    & 0.42    & 0.24    \\ \hline
MMLU                  & Acc - NLU & 50.5  & 38.66   & 34.96   & 23.1    & 39.39   & 40      & 45.85   & 57.35   \\ %\hline
BoolQ                 & Acc - NLU & 67.86 & 60.00   & 64.5    & 60.31   & 70      & 78.1    & 80      & 84      \\ %\hline
ARC Easy              & Acc - NLU & 79.97 & 60.14   & 54      & 38.8    & 59.76   & 61.24   & 71.55   & 91.16   \\ %\hline
Arc Challenge         & Acc - NLU & 65.53 & 49.83   & 35.92   & 31.66   & 48.55   & 48.29   & 59.64   & 81.23   \\ %\hline
Hella Swag       & Acc - NLU & 39.9  & 39.69    & 25.37   & 25.3    & 34.7    & 30.8    & 35.5    & 54.6    \\ \hline

IndicQuest (En)                     & Score (1-5)  & 4.01    & 3.94    & 3.75    & 3.78    & 4.1     & 4.07    & 4.2     & 4.4     \\ %\hline
IndicQuest (Hi)                     & Score (1-5)  & 4.15    & 2.72    & 3.1     & 3.18    & 3.58    & 3.6     & 4.02    & 4.23    \\ %\hline
SubjectiveEval (Hi)                & Score (1-5)  & 4.35    & 1.64    & 2.24    & 1.75    & 3.66    & 2.97    & 3.98    & 4.5     \\ %\hline

\bottomrule
\end{tabular}
        }
    \caption{Performance metrics for various instruct models across different Hindi tasks. The results are zero-shot unless otherwise specified.}
    \label{tab:model_performance_chat}
\end{table*}

\begin{table}[h!]
    \centering
    \resizebox{\columnwidth}{!}{
        \begin{tabular}{lccc}
\toprule
\textbf{Task}            & \textbf{Nemotron-Mini-Hindi-4B-Base} & \textbf{Nemotron-Mini-4B-Base} & \textbf{Gemma-2 2b} \\ \hline
\midrule
MMLU (5)                     & 56.37                          & 58.60        &   51.3         \\ \hline
arc\_challenge (25)            & 46.08                          & 50.90      &   55.4            \\ \hline
% gsm8k                    & 10.01                          & 24.10          &          \\ \hline
hellaswag (10)                & 74.64                          & 75.00       &     73          \\ \hline
truthfulqa\_mc2 (0)          & 41.05                          & 42.72        &     -      \\ \hline
winogrande (5)               & 70.09                          & 74.00        &    70.9         \\ \hline
xlsum\_english (3)           & 29.71                          & 29.62        &    -       \\ %\hline
\bottomrule
\end{tabular}
        }
    \caption{Performance of base models on English Benchmarks}
    \label{tab:model_performance_english}
\end{table}

\begin{table}[h!]
    \centering
    \resizebox{\columnwidth}{!}{
        \begin{tabular}{lccc}
\toprule
\textbf{Model Setting} & \textbf{SubjectiveEval} & \textbf{IndicQuest (Hi)} & \textbf{IndicQuest (En)} \\
\midrule
\textbf{Nemotron-Mini-4B-Base} & & & \\
\quad SFT (En) + DPO (En) & 1.92 & 2.66 & 3.89 \\
\quad SFT (En) + DPO (En + Hi) & 1.88 & 2.80 & 3.87 \\
\quad SFT (En + Hi) + DPO (En) & 2.73 & 3.20 & 3.86 \\
\quad SFT (En + Hi) + DPO (En + Hi) & 2.51 & 3.14 & 3.88 \\
\midrule
\textbf{Nemotron-Mini-Hindi-4B-Base} & & & \\
\quad SFT (En) + DPO (En) & 3.81 & 4.12 & 4.02 \\
\quad SFT (En) + DPO (En + Hi) & \textbf{4.3} & 4.10 & \textbf{4.03} \\
\quad SFT (En + Hi) + DPO (En) & 4.28 & 4.06 & 4.02 \\
\quad SFT (En + Hi) + DPO (En + Hi) & 4.25 & \textbf{4.13} & 4.04 \\
\bottomrule
        \end{tabular}
    }
    \caption{Ablation study of post-training configurations analyzing the impact of Hindi pretraining on SubjectiveEval (Chat capability) and IndicQuest (Factual accuracy) tasks.}
    \label{tab:ablation_hindi}
\end{table}

\subsection{Model Alignment}
The first alignment stage is Supervised Fine-Tuning (SFT). We use a general SFT corpus with approximately 200k examples, comprising various tasks as outlined in \cite{adler2024nemotron}. The model is trained for one epoch with a global batch size of 1024 and a learning rate in the range of [5e-6, 9e-7], using cosine annealing. Due to the lack of a high-quality Hindi SFT corpus, we leverage English-only data for SFT. We also experimented with translated English data (filtered using back-translation-based methods) for SFT, but did not observe any improvements with this addition. We found that using the English-only SFT corpus enhances instruction-following capabilities in Hindi, highlighting the cross-lingual transferability of these skills. For the ablation study, we use approximately 70k high-quality Hindi examples selected from a larger pool of 200k translated samples. These Hindi instances are derived by translating the original English data, with noisy or low-quality translations filtered out using a back-translation-based filtering approach.

After SFT stage, the model undergoes a preference-tuning phase, where it learns from triplets consisting of a prompt, a preferred response, and a rejected response. In this stage, we apply the Direct Preference Optimization (DPO) \cite{rafailov2024direct} algorithm, which trains the policy network to maximize the reward difference between the preferred and rejected responses. We train the model for one epoch with a global batch size of 512 and a learning rate in the range of [9e-6, 9e-7], utilizing cosine annealing. For the DPO stage, we use approximately 200k English samples and 60k synthetic Hindi samples. The synthetic Hindi samples were created by translating the English samples and then filtered using back-translation methods. We observe that incorporating synthetic Hindi samples during this stage improves the overall performance of the model. The aligned model is referred to as Nemotron-Mini-Hindi-4B-Instruct. Both the SFT and DPO stages are carried out using Nemo Aligner \cite{shen2024nemo} and 64 Nvidia A100 GPUs.

\begin{figure}[h]  
    \centering
    \includegraphics[width=\columnwidth]{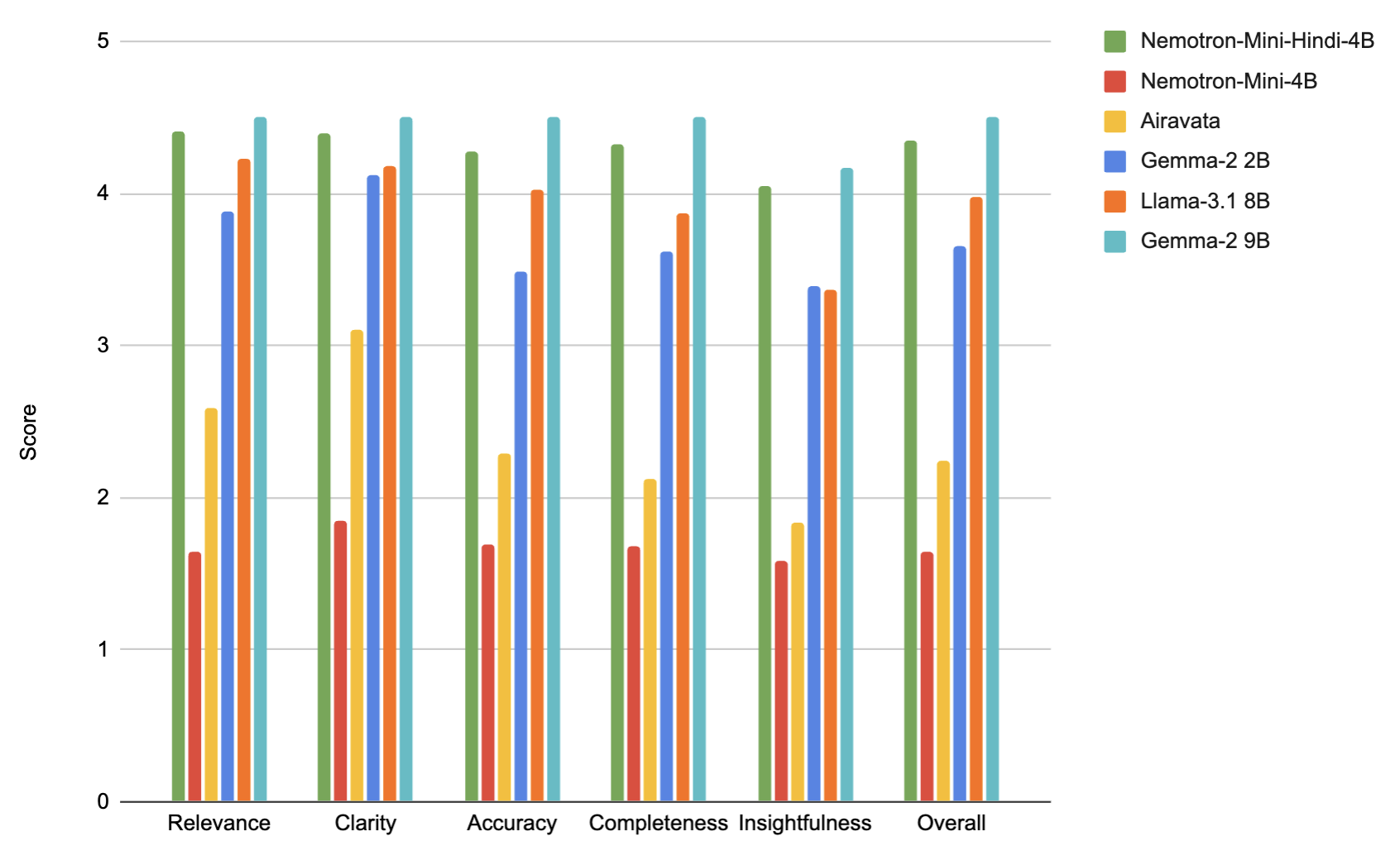}  
    \caption{Comparison of different instruct models on various parameters using SubjectiveEval.}
    \label{fig:subj_all}
\end{figure}

\begin{figure}[h]  
    \centering
    \includegraphics[width=\columnwidth]{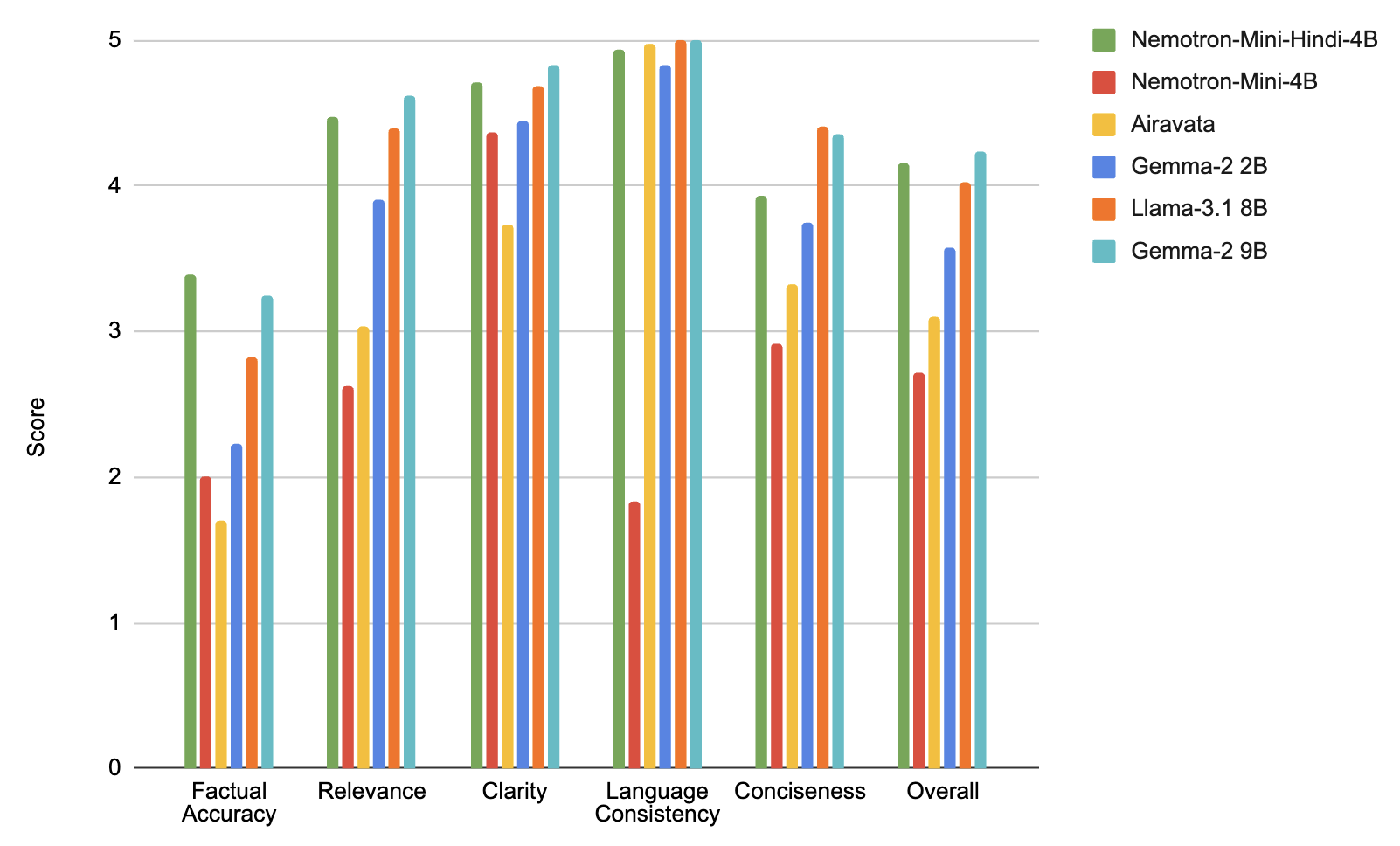}  
    \caption{Comparison of different instruct models on various parameters using IndicQuest-Hi.}
    \label{fig:quest_all}
\end{figure}

\begin{figure}[h]  
    \centering
    \includegraphics[width=\columnwidth]{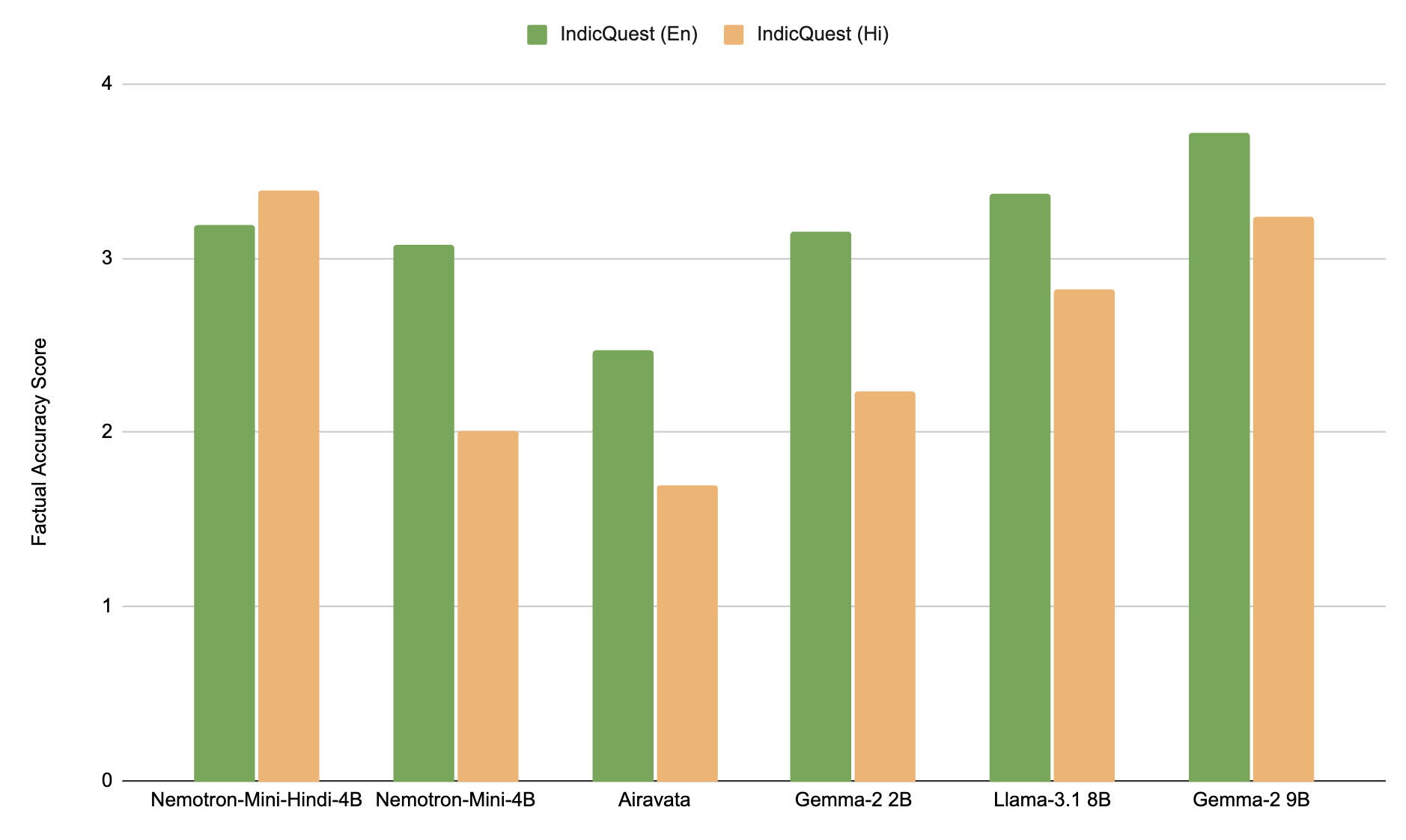}  
    \caption{Comparison of different instruct models on Factuality score of IndicQuest. The ground truth answers from IndicQuest are provided as a reference to GPT4 for better scoring. The Nemotron-Mini-Hindi-4B provides comparable scores for Hindi and English whereas other models provide better factuality for English.}
    \label{fig:factual_all}
\end{figure}

\begin{figure*}[h]  
    \centering
    \includegraphics[scale=0.14]{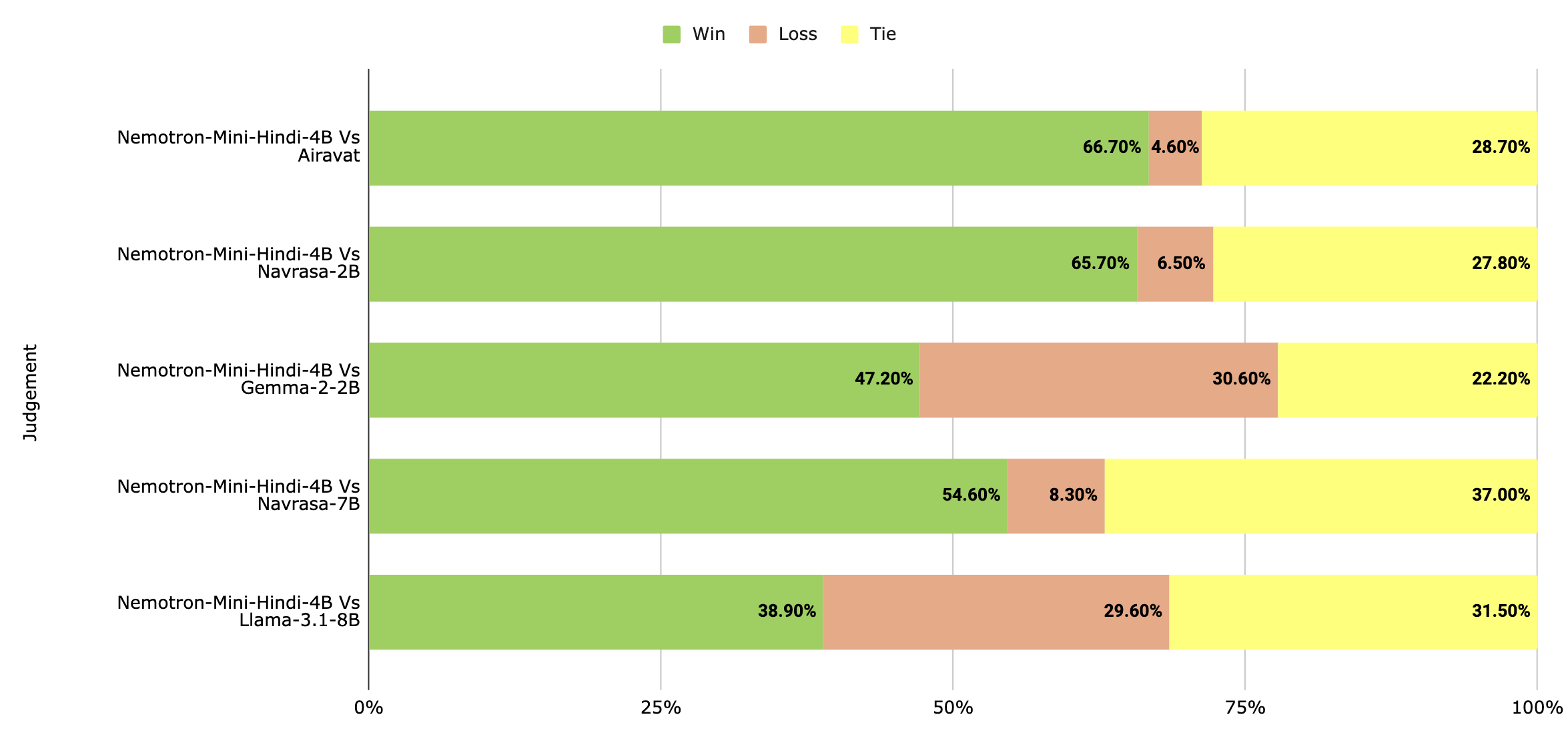}  
    \caption{Results of human evaluation on translated MT-Bench. A win indicates Nemotron-Mini-Hindi-4B model is preferred.}
    \label{fig:manual_mt_bench}
\end{figure*}

\subsection{Evaluation Datasets}
We evaluate Nemotron-Mini-Hindi-4B and other multilingual LLMs using both native Hindi benchmarks and translated English benchmarks. The native benchmarks include tasks from IndicXTREME, IndicNLG, and IndicQuest, while the translated English benchmarks include popular datasets like MMLU and Hellaswag. Additionally, we curate an open-ended QnA dataset termed SubjectiveEval to assess the model's generation capabilities in the Hindi language. Human evaluation is also conducted using the translated MT-Bench dataset. %The dataset details are provided in Appendix.

\begin{itemize}
    \item \textbf{IndicXTREME}: The benchmark consists of different Natural Language Understanding (NLU) tasks in Indic languages \cite{doddapaneni2023towards}. We consider different tasks like IndicSentiment, IndicCopa, IndicXNLI, and IndicXParaphrase.
    \item \textbf{IndicNLG}: The IndicNLG benchmark \cite{kumar2022indicnlg} consists of various tasks for evaluating the generation capabilities of the model. We consider IndicHeadline, IndicWikiBio, and IndicQA covering text summarization and question-answering tasks.
    \item \textbf{IndicQuest}: IndicQuest \cite{rohera2024l3cube} is a gold-standard fact-based question-answering benchmark designed to evaluate multilingual language models ability to capture regional knowledge across various Indic languages. It focuses on factual questions related to India in domains such as Literature, History, Geography, Politics, and Economics. The dataset is available in English as well as several Indic languages, including Hindi, allowing for language-specific evaluations. For LLM-as-a-judge evaluation, the ground truth facts are passed to the evaluator LLM as a reference.
    \item \textbf{SubjectiveEval}:
    % This in-house Hindi evaluation dataset features open-ended questions across various Indian domains, mathematics, and thinking ability. It offers broader coverage compared to the fact-based questions in IndicQuest.
    This in-house Hindi evaluation dataset features open-ended questions across various Indian domains, including History, Geography, Agriculture, Food, Culture, Religion, Science and Technology, Mathematics, and Thinking Ability. It offers broader coverage compared to the fact-based questions in IndicQuest. It assesses a model’s understanding, generative capabilities, coherence, and insightfulness. Questions include 'what', 'how', and 'why' types, varying from brief one-word answers to detailed explanations. The dataset also tests analytical and problem-solving skills with hypothetical scenarios. Model responses are evaluated using an LLM as a judge.
    
    \item \textbf{Translated English Benchmarks}:
    We use translated versions of popular benchmarks for exhaustive evaluation of our models. The benchmarks include MMLU, Hella Swag, BoolQ, Arc-Easy, and Arc-Challenge.
    \item \textbf{Human Evaluation}:
    For human evaluation, we utilized a translated version of the multi-turn MT-Bench dataset \cite{zheng2023judging}. The prompts were first translated into Hindi using the Google Translate API and then manually filtered to remove problematic prompts or those relying on English-specific semantics. During evaluation, human judges conducted A/B testing, where they were presented with randomized, pair-wise model responses for comparison.
\end{itemize}

\section{Results and Discussion}
The results for the base models are shown in Table \ref{tab:model_performance_base} and Table \ref{tab:model_performance_english}. The Nemotron-Mini-Hindi-4B Base delivers state-of-the-art performance on nearly all benchmarks compared to similarly sized models. Additionally, it outperforms larger models like Gemma-2-9B and Llama-3.1-8B on more than half of the benchmarks. Hindi-specific continued pre-training significantly enhances the model's performance on Hindi tasks compared to the base Nemotron-Mini-4B model. There is some degradation on English benchmarks, though the results remain competitive. This underscores the importance of dual-language continued pre-training.

We observe similar results with the instruct model on IndicXTREME, IndicNLG, and translated English benchmarks. The results are presented in Table \ref{tab:model_performance_chat}. The instruct model is also evaluated using LLM-as-a-judge on IndicQuest and SubjectiveEval. On these benchmarks, we see improvements in both English and Hindi compared to the Nemotron-Mini-4B-Instruct model. The model outperforms all baseline models except for Gemma-2-9B. Notably, we observe improvements in the model's factuality and language consistency. These results are shown in Figure \ref{fig:subj_all}, \ref{fig:quest_all}, and \ref{fig:factual_all}. Furthermore, during human evaluations, responses from Nemotron-Mini-4B-Hindi were consistently preferred over those from other models, as shown in Figure \ref{fig:manual_mt_bench}.
%Refer Appendix for detailed plots.

Table~\ref{tab:ablation_hindi} shows an ablation study analyzing the impact of Hindi SFT and DPO on Nemotron-Mini-4B Base and the Hindi-pretrained Nemotron-Mini-Hindi-4B Base. While Hindi alignment improves the base Nemotron-Mini-4B, its performance remains significantly lower than the Hindi-pretrained model across all metrics. Notably, Nemotron-Mini-Hindi-4B with English alignment alone outperforms Nemotron-Mini-4B even with Hindi alignment, highlighting the strong benefits of Hindi pretraining. The best results are achieved with English SFT and (English + Hindi) DPO on the Hindi-pretrained model, suggesting that mixed-language alignment can enhance generalization. Overall, Hindi pretraining leads to substantial improvements not only in Hindi understanding but also in factual accuracy and English performance, demonstrating effective cross-lingual transfer.

\section{Conclusion}
We present Nemotron-Mini-Hindi-4B-Base and Nemotron-Mini-Hindi-4B-Instruct, state-of-the-art SLMs primarily designed for the Hindi language. These models have been continuously pre-trained and aligned using a combination of Hindi and English data. The Hindi corpus includes both real and synthetic data, with the synthetic data generated through translation. The models outperform similarly sized models on various Hindi benchmarks, as assessed through reference-based and LLM-as-a-judge evaluations. They also perform competitively on English benchmarks. We emphasize the importance of pre-training to reduce hallucinations and enhance the factuality of the models.

\section*{Limitations}
The model was trained on internet data that includes toxic language and biases, which means it might reproduce these biases and generate toxic responses, particularly if prompted with harmful content. It may also produce inaccurate, incomplete, or irrelevant information, potentially leading to socially undesirable outputs. The problem could be worsened if the suggested prompt template is not used.

To mitigate these issues to some extent, we have implemented safety alignment during the DPO stage to guide the model away from responding to toxic or harmful content. Additionally, we conduct safety evaluations using benchmarks such as Aegis\footnote{https://huggingface.co/datasets/nvidia/Aegis-AI-Content-Safety-Dataset-1.0} \cite{ghosh2024aegis}, Garak\footnote{https://github.com/leondz/garak} \cite{derczynski2024garak}, and Human Content red-teaming, and our findings indicate that the model's responses remain within permissible limits.

\section*{Acknowledgements}
This work would not have been possible without contributions from many people at NVIDIA. To mention a few: Asif Ahamed, Ayush Dattagupta, Umair Ahmed, Yoshi Suhara, Ameya Mahabaleshwarkar, Zijia Chen, Varun Singh, Vibhu Jawa, Saurav Muralidharan, Sharath Turuvekere Sreenivas, Marcin Chochowski, Rohit Watve, Oluwatobi Olabiyi, Mostofa Patwary, and Oleksii Kuchaiev.   

\bibliography{main}

\appendix

% \section{Example Appendix}
% \label{sec:appendix}

% This is an appendix.

\end{document}